\documentclass{article}

 \usepackage[final, nonatbib]{neurips_workshop_2019_hydro_model}

\usepackage[utf8]{inputenc} 
\usepackage[T1]{fontenc}    
\usepackage{hyperref}       
\usepackage{url}            
\usepackage{color}          
\usepackage{graphicx}       
\usepackage{booktabs}       
\usepackage{amsfonts}       
\usepackage{nicefrac}       
\usepackage{multirow}
\usepackage{booktabs}
\usepackage{tablefootnote}
\usepackage[bottom]{footmisc}
\usepackage{microtype}      
\usepackage{amssymb}            
\title{Accurate Hydrologic Modeling Using Less Information}

\author{%
  Guy Shalev\thanks{Google Research} \\
  \texttt{guysha@google.com} \\
  \And
  Ran El-Yaniv\footnotemark[1] \ \thanks{Technion – Israel Institute of Technology} \\
  \texttt{rani@cs.technion.ac.il} \\
  \And
  Daniel Klotz\thanks{LIT AI Lab \& Institute for Machine Learning, Johannes Kepler University Linz, Austria} \\
  \texttt{klotz@ml.jku.at} \\
  \AND
  Frederik Kratzert \footnotemark[3] \\
  \texttt{kratzert@ml.jku.at} \\
  \And
  Asher Metzger\footnotemark[1]\\
  \texttt{ashermetzger@google.com} \\
  \And
  Sella Nevo\footnotemark[1] \\
  \texttt{sellanevo@google.com} \\
}

\begin{document}
\maketitle

\begin{abstract}

Joint models are a common and important tool in the intersection of machine learning and the physical sciences, particularly in contexts where real-world measurements are scarce. Recent developments in rainfall-runoff modeling, one of the prime challenges in hydrology, show the value of a joint model with shared representation in this important context. However, current state-of-the-art models depend on detailed and reliable attributes characterizing each site to help the model differentiate correctly between the behavior of different sites. This dependency can present a challenge in data-poor regions. In this paper, we show that we can replace the need for such location-specific attributes with a completely data-driven learned embedding, and match previous state-of-the-art results with less information.

\end{abstract}

\section{Introduction}

The prediction of hydrologic processes is critical for utilizing and protecting against the immense impacts of water on human life. These impacts can include mitigating the effect of floods, which are responsible for thousands of fatalities and billions of dollars in economic damages annually, improving agriculture, which is responsible for the livelihood of a significant portion of humanity, and more.
Hydrologic models allow for simulating and forecasting various water flow properties (most often streamflow) based on more easily measurable inputs (most notably precipitation).

However, despite their economic and humanitarian importance, reliable hydrologic models remain a challenge in developing countries, mainly due to poor quality or unavailability of data. Our goal is to bridge this gap and enable effective hydrologic modeling at scale in data-scarce regions. In this paper we present a hydrologic model that works with partial data -- while retaining state-of-the-art performance.

Classical hydrologic models (referred to by hydrologists as ``conceptual models'') are based on equations that describe the physics of the rainfall-runoff process. Traditionally, the parameters of these models are optimized separately for each geographic location (a.k.a. ``site'' or ``basin'') using site-specific data \cite{beven2011rainfall}.
Attempts to construct a conceptual model applicable to many sites (a.k.a. ``regional'' or ``joint'' model) typically achieve significantly inferior performance \cite{mizukami2017towards,mizukami2019choice,Kratzert2019}.

Recently, Kratzert et al. \cite{Kratzert2019} used a Long Short-Term Memory (LSTM) network to create a single (regional) hydrologic model for hundreds of sites from the extensive CAMELS hydrologic dataset \cite{Newman2014, Addor2017} -- the single LSTM-based model was trained on the combined input data from hundreds of sites to predict the streamflow data in all of these basins. Their regional LSTM model -- when equipped with basin-specific attribute features (such as altitude and aridity) -- significantly outperformed all of the prominent conceptual models -- both regional models and models that were calibrated per basin. This important result proves that a joint machine learning model approach is promising compared to the single site approach that suffers from inherent data shortage.


When considering hydrologic modeling in developing world regions, data scarcity is a major concern. Shorter or irregular discharge records prevent the creation of accurate site-specific models, emphasizing the necessity of the joint modeling approach which allows the transfer knowledge between the tasks of predicting for different basins.
The state-of-the-art results achieved by Kratzert et al.'s \cite{Kratzert2019} joint model strongly depend on site specific static attributes (e.g., soil characteristics), that help the model to effectively utilize the abundance of diverse data provided from aggregating hundreds of sites, while retaining the ability to differentiate between the hydrologic response of different locations.
Unfortunately, their approach is not directly applicable in the developing world, where some of these basin attributes are of poor quality and require significant efforts in curating the relevant datasets \footnote{Datasets similar to CAMELS have only been produced for few other regions - Chile, Great Britain and Australia (e.g., \cite{alvarez2018camels})}. Therefore, obtaining good performance without relying on these attributes is an important step towards scalable and accurate hydrologic modeling in developing countries. 

In this paper we present a joint hydrologic model that does not rely on any site static features. We replace these features by a site embedding layer, similar to the word-embedding technique used in language models \cite{mikolov2013distributed}. Our empirical study over the CAMELS dataset validates this approach and clearly demonstrates that the proposed model replicates state-of-the-art performance with less information. This success can be interpreted as our model's ability to learn the specific hydrologic response of each site directly from the meteorologic and streamflow dynamics at comparable effectiveness to its ability to learn that response from static basin attributes (see a detailed discussion in Section \ref{restuls-discussion}).

\section{Model Architecture and Optimization}
\label{architecture}
We train a single (regional) LSTM on the combined data of hundreds of sites to predict the mean discharge of a single day, given the history of the meteorological input features of the previous 270 days. To allow a fair comparison, our model setting is similar to the LSTM setup described in detail in \cite{Kratzert2019}: a single layer, 256 memory cells and a single fully connected layer with a droupout rate of 0.4.

We consider two types of input features:
\begin{itemize}
    \item Meteorologic time series data at a daily resolution (such as daily basin-averaged precipitation and temperature). 
    \item Static features (e.g., basin size, fraction of sand in the soil), representing attributes that are constant for each site.
\end{itemize}

For every timestep $t=1,...,270$ we feed the model with dynamic data from day $t$, and produce a prediction of the daily discharge mean at day $t=270$.

In experiments where static features are used, they are concatenated to the dynamic inputs at each time step.
In experiments where our proposed site embedding is used, the embedding is a vector $v_i \in \mathbb{R}^k$ (for some hyperparameter $k$), that is learned during training for each site $i=1, ...,n$. The embedding vector is concatenated at every time step, similarly to the static features it aims to substitute. In the reported results, we use $k=20$ as the embedding dimension.

All input features (both static and dynamic) and labels were standardized (zero mean, unit variance).

Our loss function is the basin averaged Nash-Sutcliffe Efficiency (see Section \ref{metrics}), with a constant term in the denominator to allow robustness of the optimization to catchments with very low flow-variance. A detailed discussion of this loss function is available in \cite{Kratzert2019}.

\section{The NCAR CAMELS Dataset}
\label{camels-data}
As mentioned earlier, to be able to benchmark our model against all models presented in the work of Kratzert et al. \cite{Kratzert2019}, we aspired to work in a setting as similar as possible to theirs in terms of features, hyperparameters, etc. 
We therefore work with the CAMELS dataset, which consists of 671 basins across the Contiguous United States (CONUS). We use five dynamic features provided in the dataset, which are the daily, basin-averaged Maurer meteorologic forcings \cite{wood2002long}. These include: (i) precipitation, (ii) minimum air temperature, (iii) maximum air temperature, (iv) average short-wave radiation and (v) vapor pressure. As static features, we also utilize the same 27 basin attributes listed in the appendix of \cite{Kratzert2019}. The time periods for all experiments are compatible with the benchmarks: Oct. 1999 to Sep. 2008 for training (9 full years) and Oct. 1989 to Sep. 1999 for evaluation (10 full years). The daily mean discharge labels are measurements published by the United States Geological Survey (USGS). Our model is trained and evaluated on 528 sites out of the 531 that were used in Kratzert's paper -- three sites were excluded due to data parsing issues before modeling.

\section{Performance Metrics}
\label{metrics}
The Nash-Sutcliffe Efficiency (NSE, Nash \& Sutcliffe, 1970) \cite{nash1970river} is the most common metric in hydrology for the evaluation of rainfall-runoff models for single sites. It is defined as the $R^2$ between the simulated and observed discharge:

$$ 1 - \frac{\sum_{t=1}^T (Q_m[t] - Q_o[t])^2}{\sum_{t=1}^T (Q_o[t] - \bar{Q_o})^2}$$

Where for every example $t=1,2,\dots,T$, we denote $Q_m[t]$ the modeled discharge and $Q_o[t]$ the observed discharge at time $t$. $\bar{Q_o}$ is the mean observed discharge. This is equivalent to the MSE normalized by the variance, and then subtracted from 1. Note that possible ranges of the metric are $(-\inf, 1]$, with 1 obtained for perfect simulation and 0 for the model that constantly predicts the mean observed discharge for the site. 

We compute the NSE for each of the sites on the joint model's predictions. We then aggregate them into the 3 central metrics benchmarked on the CAMELS dataset: median NSE, mean NSE, and number of sites with negative NSE score (i.e., worse than predicting the mean).

\section{Results and Discussion}
\label{restuls-discussion}

The main results of our experiments are presented in Table \ref{camels-results} along with the  results obtained in \cite{Kratzert2019}  that are relevant for comparison.

\begin{table}[ht]
  \caption{Evaluation of the models on the CAMELS test dataset}
  \label{camels-results}
  \centering
  \begin{tabular}{lccc}
    \toprule
    \multirow{2}{*}{Model} & \multicolumn{2}{c}{NSE} & No. of basins \\
    & mean & median & with NSE $\leq 0$ \\
    \midrule
    LSTM w/o static inputs\tablefootnote{Results reported in \cite{Kratzert2019}}                    & 0.39  & 0.59  & 28   \\
    LSTM with static inputs\footnotemark[5]                   & 0.69  & 0.73  & 2  \\
    LSTM w/o static inputs, with embedding    & 0.69  & 0.73  & 1   \\
    LSTM with static inputs, with embedding   & 0.70  & 0.73 & 2   \\
    \\
    \bottomrule
  \end{tabular}
\end{table}

From the results presented in Table \ref{camels-results} we conclude that replacing static attributes with site embedding is a viable approach for regional modeling, achieving almost identical performance on the measured metrics. One potential explanation for this convergence is that the dynamics of the temporal features (and labels) provided to the model are rich and indicative enough to enable the model to fully identify the relevant information within the static features -- at least when the historical record is long enough (9 years of daily data in our case).

The last row in Table \ref{camels-results} is consistent with this hypothesis. Despite both the static inputs and the site embedding providing a significant improvement over the model without static inputs, providing both of these tools to the model does not improve its performance.

Note that both the static features and the site embedding can be useful tools in addressing different types of in-availability of data. Clearly, in the theoretical scenario of infinite historical training data\footnote{This also requires the assumption that the training data and test data are identically distributed, an assumption we know is never perfectly fulfilled in a real-world physical system.}, one would expect the embedding to perform better than static attributes -- the provided attributes are sometimes noisy estimations of the actual values they represent (e.g. due to measuring errors), and also do not incorporate all the site-specific information that can be helpful for the model and available from the data (e.g., the mean and variance of the discharge values). At the other extreme, when tasked with predicting discharge at a site with no discharge measurements at all (prediction in ``ungauged basins'' \cite{bloschl2013runoff}), the embedding approach is inapplicable, but the basin attributes are still useful -- as explored in \cite{kratzert2019prediction}.
We therefore see both approaches as being of significant and complementary value. An interesting question for future research would be how many samples (per site) are necessary to learn a good embedding representation, clarifying the constraints of each approach.


\section{Conclusions and Future Work}
In this paper we have shown that state-of-the-art results can be matched without relying on additional information which can be of poor quality or difficult to curate in many regions.

We view these results as an important milestone in the path to real world, scalable flood forecasting applications in the developing world, as detailed in Nevo et al. \cite{Nevo2019ML-FF-Scale}. We believe this goal is critical not only from a scientific point of view, but also from a humanitarian perspective. The vast majority of the thousands of annual flood-related fatalities occur in developing countries, where data quality and availability are a critical issue.

A natural next step is to explore the effectiveness of this approach when applied directly to severely affected developing countries such as India, Bangladesh and more. This type of research has significant challenges -- including access to the data necessary for both prediction and evaluation -- but is critical in converting these and similar results into actual real-world impact.

There are several other research directions that arise from our results. Possible future work could focus on examining the interpretability of the embedding layer -- for example, predicting various basin attributes (e.g., altitude or soil characteristics) directly from the embedding, or analyzing the clustering of sites with similar embeddings with relation to other clustering works on the CAMELS sites \cite{camels2019cluster,Kratzert2019}.

We believe the flexibility of joint-model, data-driven approaches can help overcome many challenges that arise from data constraints. Some of these challenges are not effectively addressed by classic tools, while others are well-handled by conceptual models and would be of value to import to machine learning based models. Examples of these include:
\begin{itemize}
    \item Aggregation of many meteorological inputs. There exist a significant number of precipitation products, which clearly contain more information than any one of them separately. Machine learning based hydrologic models can utilize several products simultaneously without the need to explicitly combine them into one precipitation estimate.
    \item Using non-standard labels such as water level. Reliable water level measurements are much more commonly available than discharge measurements, yet conceptual models do not model these well because they do not follow simple conservation laws. Data-driven approaches are very well-placed to implicitly identify the correlations between discharge and water level, and can therefore utilize these better both as input and as labels.
    \item Utilising upstream measurements and forecasts. Classic approaches use routing models to utilize upstream measurement in a fairly straightforward manner, but incorporating these into machine learning models raises many interesting architectural questions.
    \item Severe data scarcity. Dealing with small training sets is a core area of research in machine learning, and is extremely relevant to this space where the real-world impact of a model tends to correlate strongly with the lack of data available (due to developing countries both lacking data collection infrastructure, while depending more on these models for basic needs in safety and agriculture).
    \item Utilization of site attributes when those are available. These attributes are hard to incorporate efficiently into conceptual models.
    
\end{itemize}

The above present both challenges and opportunities for the hydrology and machine learning communities, and we hope further collaborations between these two disciplines will produce significant results, both academically and operationally.

\bibliography{neurips_workshop_2019_hydro_model}
\bibliographystyle{plain}

\end{document}